\documentclass{article}

\PassOptionsToPackage{numbers, compress}{natbib}
\usepackage{amsmath}%添加的

\usepackage[final]{neurips_2025}

\usepackage[utf8]{inputenc} % allow utf-8 input
\usepackage[T1]{fontenc}    % use 8-bit T1 fonts
\usepackage{hyperref}       % hyperlinks
\usepackage{url}            % simple URL typesetting
\usepackage{booktabs}       % professional-quality tables
\usepackage{amsfonts}       % blackboard math symbols
\usepackage{nicefrac}       % compact symbols for 1/2, etc.
\usepackage{microtype}      % microtypography
\usepackage{xcolor}         % colors
\usepackage{graphicx}
\usepackage{multirow} %添加的
\title{UGG-ReID: Uncertainty-Guided Graph Model for Multi-Modal Object Re-Identification}
\author{%
    Xixi Wan$^1$\;\;\;  Aihua Zheng$^{1}$\thanks{Corresponding Author}\;\;\;   Bo Jiang$^{2*}$\;\;\; Beibei Wang$^{2}$\;\;\;Chenglong Li$^1$\;\;\; Jin Tang$^{2}$\\
  $^1$Information Materials and Intelligent Sensing Laboratory of Anhui Province, \\ School of Artificial Intelligence, Anhui University, Hefei, China\\ 
  $^2$Anhui Provincial Key Laboratory of Multimodal Cognitive Computation, \\ School of Computer Science and Technology, Anhui University, Hefei, China\\ 
  \texttt{$^1$ahzheng214@foxmail.com, $^2$jiangbo@ahu.edu.cn}\\ 
}
\begin{document}

\maketitle

\begin{abstract}
Multi-modal object Re-IDentification (ReID) has gained considerable attention with the goal of retrieving specific targets across cameras using heterogeneous visual data sources.
At present, multi-modal object ReID faces two core challenges: (1) learning robust features under fine-grained local noise caused by occlusion, frame loss, and other disruptions; and (2) effectively integrating heterogeneous modalities to enhance multi-modal representation.
To address the above challenges, we propose a robust approach named Uncertainty-Guided Graph model for multi-modal object ReID (UGG-ReID). UGG-ReID is designed to mitigate noise interference and facilitate effective multi-modal fusion by \textbf{estimating both local and sample-level aleatoric uncertainty} and \textbf{explicitly modeling their dependencies}.
Specifically, we first propose the Gaussian patch-graph representation model that leverages uncertainty to quantify fine-grained local cues and capture their structural relationships. This process boosts the expressiveness of modal-specific information, ensuring that the generated embeddings are both more informative and robust. Subsequently, we design an uncertainty-guided mixture of experts strategy that dynamically routes samples to experts exhibiting low uncertainty. This strategy effectively suppresses noise-induced instability, leading to enhanced robustness. Meanwhile, we design an uncertainty-guided routing to strengthen the multi-modal interaction, improving the performance.
UGG-ReID is comprehensively evaluated on five representative multi-modal object ReID datasets, encompassing diverse spectral modalities. Experimental results show that the proposed method achieves excellent performance on all datasets and is significantly better than current methods in terms of noise immunity. \textbf{Our code is available at https://github.com/wanxixi11/UGG-ReID.}

\end{abstract}

\section{Introduction}
Multi-modal data has emerged as a significant trend in artificial intelligence~\cite{10657706, 10655615, 10.1145/3696410.3714606, 10841969}. Especially driven by large model technology, more and more applications have begun to utilize multi-modal information for comprehensive analysis~\cite{mckinzie2024mm1, 10655344, 10.5555/3692070.3694257}. Multi-modal object Re-IDentification (ReID)~\cite{Li2020MultiSpectralVR, Zheng2021RobustMP, ZHENG2023101901, ZHENG2025102800}, as a cutting-edge direction of this research, not only broadens the application boundaries of traditional object ReID~\cite{9011001,10807364,s44267-023-00032-9}, but also effectively makes up for some limitations in cross-modal object ReID~\cite{Dai2018CrossModalityPR,10.1145/3343031.3351006, He_2021_ICCV}.

\begin{figure}
  \centering
  \includegraphics[width=1\linewidth]{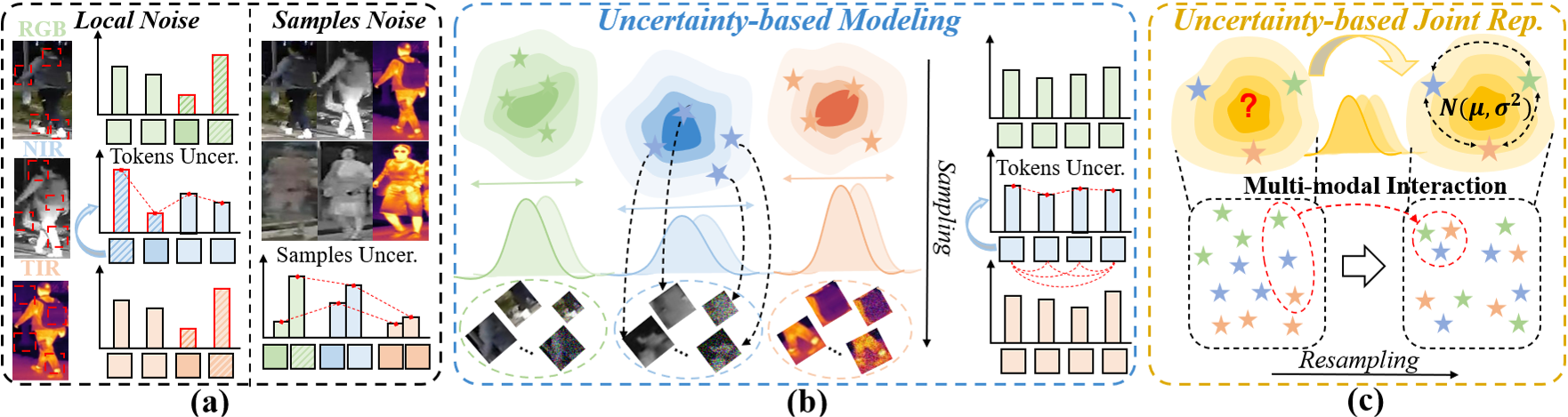} 
  \label{motivation}
  \vspace{-1.5em}
  \caption{Modeling aleatoric uncertainty in multi-modal ReID. (a) The challenges in multi-modal object ReID. (b) Local uncertainty modeling for modality-specific refinement. (c) Joint uncertainty modeling in samples and modalities for improved fusion.}
  \label{overall}
  \vspace{-1.5em}
\end{figure}

Recently, researchers have extensively explored multi-modal feature fusion and matching strategies to bridge the representation gap between different modalities for object ReID~\cite{ Zheng2021RobustMP, Wang2024HeterogeneousTT, Wang2024MambaPro, Yang2025TIENet}.
For example, 
Zheng \emph{et al.}~\cite{Zheng2021RobustMP} propose a progressive fusion method to achieve effective fusion of multi-modal data.
Wang \emph{et al.}~\cite{Wang2024HeterogeneousTT} propose HTT that exploits the relationship on unseen test data between heterogeneous modalities to improve performance. 
On the other hand, existing methods~\cite{Wang2023TOPReIDMO, 10654953, wang2025IDEA, Zhang25PromptMA} have begun to pay attention to the effect of modal noise on the discriminative properties of local regions, such as the introduction of local region alignment, noise suppression module or sample reconstruction, to enhance the robustness and improve the performance of the model. Among these,
Zhang \emph{et al.}~\cite{10654953} propose EDITOR to suppress interference from background information and promote feature learn. 
Wang \emph{et al.}~\cite{wang2025IDEA} propose utilizing CDA to focus on localized regions of the discriminatory.
% Zhang \emph{et al.}~\cite{Zhang25PromptMA} propose PromptMA to establish effective connections among different complementary modal information.
We can observe that some methods usually assume that the quality and representation of each modal data are balanced and stable, and ignore the local and sample noise disturbances within the modality caused by factors such as occlusion, low resolution~\cite{Zheng2021RobustMP, Wang2024HeterogeneousTT}. 
Although existing works~\cite{10654953, wang2025IDEA, YangTwinNoisy, YangRobust, Qin2023NoisyCorrespondenceLF} have achieved some success in mitigating the impact of noise on network performance, there are still significant shortcomings in their approaches when confronted with inconsistent local noise patterns and samples with different noise intensities. As depicted in Fig.~\ref{motivation} (a), the lack of fine-grained local noise learn and sample-level noise handling compromises the robustness of multi-modal fusion, which increases multi-modal uncertainty and impairs overall model performance.

Therefore, to solve the above problems, we provide a robust approach named Uncertainty-Guided Graph model for multi-modal object ReID (UGG-ReID). The proposed method quantifies local aleatoric uncertainty and models their structural dependencies. Following this, leveraging per-sample uncertainty guides feature refinement, promoting more reliable modality interaction.
To be specific, we first propose the Gaussian Patch-Graph Representation (GPGR), which encourages fine-grained local features to conform to Gaussian distributions. Meanwhile, a Gaussian patch graph is constructed to explicitly model the dependencies among these local features, thereby capturing fine-grained consistency information. As illustrated in Fig.~\ref{motivation} (b), the uncertainties of local tokens are effectively evaluated and explored via uncertainty-based modeling.
Second, we design an Uncertainty-Guided Mixture of Experts (UGMoE) strategy, which dynamically assigns samples to different experts based on their uncertainty levels. Besides, this strategy incorporates an uncertainty-guided routing mechanism to enhance multi-modal interaction. As shown in Fig.~\ref{motivation} (c), multi-modal data is jointly learned to improve the overall performance based on uncertainty. The entire framework is trained in an end-to-end manner. Through extensive experimentation on five public multi-modal object ReID datasets, our method not only achieves competitive performance but also consistently outperforms prior methods in noisy scenarios, validating its effectiveness and robustness.

\label{Method}
In summary, the key contributions of this method are outlined as follows:
\begin{itemize}
\item[$\bullet$] \noindent We propose the  Uncertainty-Guided Graph model for multi-modal object ReID (UGG-ReID). UGG-ReID first enhances features within each modality by considering the distribution of global and local cues and then leveraging experts' interaction to jointly capture complementary information among multi-modal data, improving model robustness and performance. 
\item[$\bullet$] \noindent We design a Gaussian Patch-Graph Representation (GPGR) to quantify aleatoric uncertainties for global and local features while modeling their relationships. GPGR can further alleviate the impact of noisy data and effectively reinforce modal-specific information. 
To our knowledge, this is the first work that leverages uncertainty to \textbf{quantify fine-grained local details} and \textbf{explicitly model their dependencies} in multi-modal data.
%dependencies
\item[$\bullet$] \noindent
We introduce the Uncertainty-Guided Mixture of Experts (UGMoE) strategy, which makes different samples select experts based on the uncertainty and also utilizes an uncertainty-guided routing mechanism to strengthen the interaction between multi-modal features, effectively promoting modal collaboration.
% \item[$\bullet$] \noindent 
% Through extensive experimentation on five public multi-modal object ReID datasets, our method not only achieves competitive performance but also consistently outperforms prior methods in noisy scenarios, validating its effectiveness and robustness.
\end{itemize}
\section{Related Works}
\subsection{Multi-Modal Object ReID}
The existing multi-modal ReID methods can be summarized into two types: one focuses on feature fusion and integration between modalities, aiming to alleviate the semantic representation differences between different modalities~\cite{Wang2024MambaPro, Zhang25PromptMA, li2025icpl, Yang25TIENet}. 
For instance,
Yang \emph{et al.}~\cite{Yang25TIENet} propose the tri-interaction enhancement network (TIENet). This method applies spatial-frequency interaction to enhance feature extraction and multi-modal fusion.
Wang \emph{et al.}~\cite{Wang2024MambaPro} propose a novel method
called MambaPro, which utilizes mamba aggregation to fuse the information of multi-modal Object ReID. 
Zhang \emph{et al.}~\cite{Zhang25PromptMA} propose PromptMA to establish effective connections among different multi-modal information.
The other type focuses on the local noise interference within the modality and improves the overall recognition performance by enhancing the local region discrimination~\cite{ZHENG2023101901, Wang2023TOPReIDMO, 10654953, wang2025IDEA}. 
Zheng \emph{et al.}~\cite{ZHENG2023101901} proposes CCNet, which seeks to simultaneously mitigate identification uncertainty due to modal variability and sample appearance changes by jointly modeling multi-modal heterogeneity and intraclass perturbations under view angle and lighting changes.
Zhang \emph{et al.}~\cite{10654953} propose EDITOR that uses 
Spatial-Frequency Token Selection (SFTS) module to select diverse tokens and suppress the effects of background interference. 
Wang \emph{et al.}~\cite{wang2025IDEA} propose the Inverted text with cooperative DEformable Aggregation (IDEA) framework to solve noise interference, enhancing feature robustness.
However, the existing methods generally lack explicit modeling of fine-grained local information quality and sample-level uncertainty, which makes it difficult for the model to effectively perceive and suppress local noise in the face of complex modal degradation or multi-source noise interference, which limits its robustness and generalization ability.

\subsection{Uncertainty in Multi-Modal Learning}
Multi-modal learning faces the challenge of uncertainty from the data layer and the model layer~\cite{10657706, 10205473, 10.1145/3664647.3680949, 10737387, 10688376, ZHANG2025110993, s44267-025-00078-x}. The former is embodied in aleatoric uncertainty, which arises from indelible perceived noise; The latter manifests as an epistemic uncertainty due to the limitations of the model's capabilities. In practice, aleatoric uncertainty is more common and has a direct impact on model performance. To this end, many works in recent years have focused on the introduction of uncertainty mechanisms to multi-modal learning to effectively identify and suppress unreliable information, improving the performance of the model. 
Ji \emph{et al.}~\cite{10205473} propose the Probability Distribution Encoder (PDE) module to aggregate all modalities into a probability distribution, framing the uncertainty and optimizing multi-modal representations.
Gao \emph{et al.}~\cite{10657706} propose quantifying the intrinsic aleatoric uncertainty of single modality to enhance multi-modal features.
Zhang \emph{et al.}~\cite{ZHANG2025110993} propose UMLMC  that employs uncertainty-guided meta-learning to mitigate feature-level bias.
Therefore, reasonable modeling and quantification of the uncertainty in multi-modal data can not only improve the robustness and generalization ability of multi-modal models but also provide more reliable decision support for practical applications.
% \subsection{Mixture of Experts}
\section{Methodology}
\label{Method}
\begin{figure}
  \centering
  \includegraphics[width=1\linewidth]{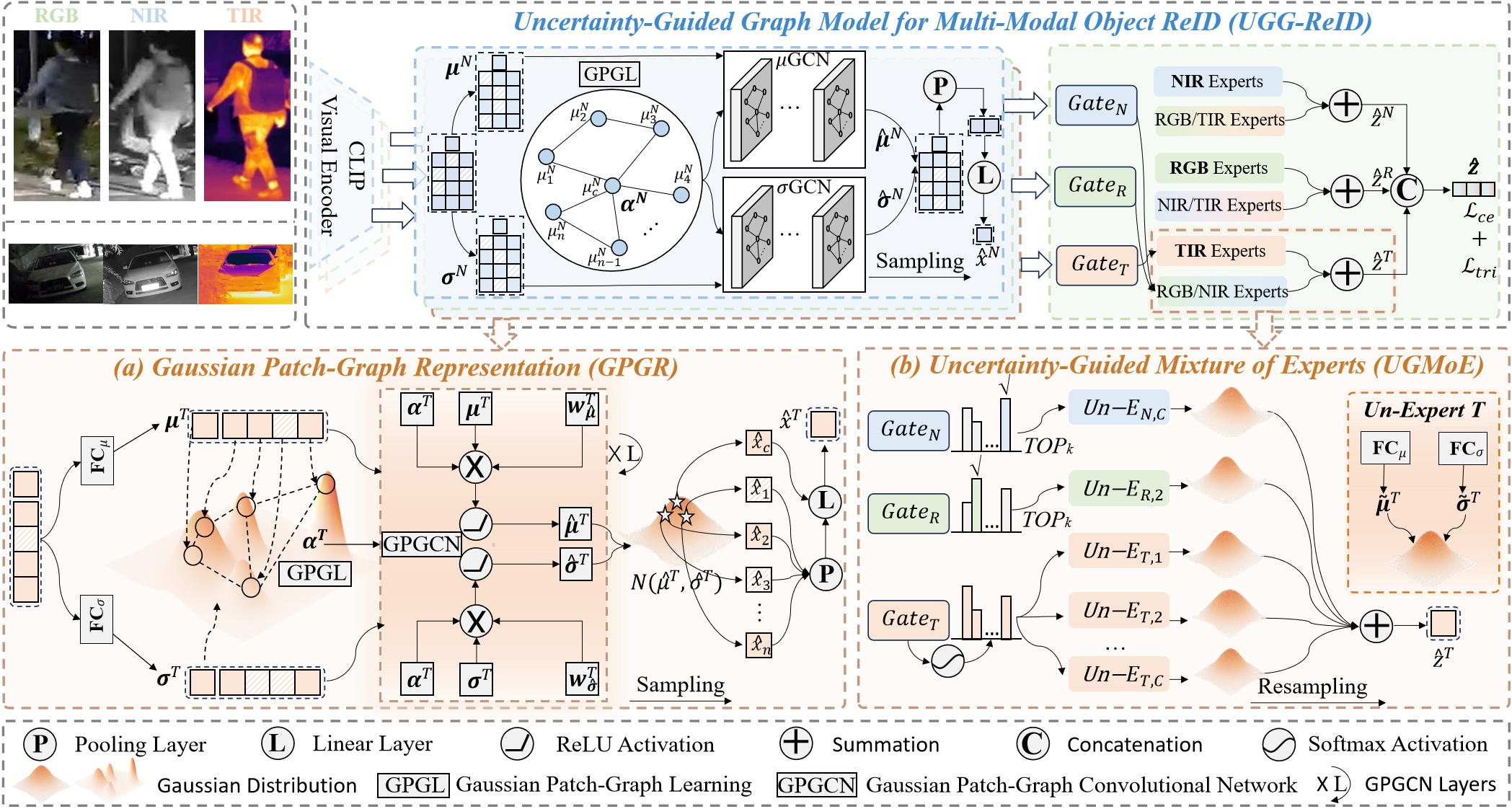} 
  \vspace{-1.5em}
  \caption{The overall framework of the proposed Uncertainty-Guided multi-modal object ReID (UGG-ReID), which is composed of two main components: Gaussian Patch-Graph Representation (GPGR) and Uncertainty-Guided Mixture of Experts (UGMoE).
  }
  % The overall framework of Uncertainty-Guided multi-modal object ReID (UGG-ReID), which consists of two parts, A and B
  \label{overall}
  \vspace{-1em}
\end{figure}

We propose a novel Uncertainty-Guided multi-modal object ReID (UGG-ReID) framework, as shown in Fig.~\ref{overall}. Specifically, the proposed UGG-ReID first designs the Gaussian Patch-Graph Representation (GPGR) model as Fig.~\ref{overall} (a), which quantifies both global and local uncertainties and models the dependencies between them. This enables richer feature representations for each modality and yields semantically more consistent embeddings. Subsequently, the Uncertainty-Guided Mixture of Experts (UGMoE) strategy makes samples select experts of low uncertainty, promoting multi-modal interactions while mitigating the propagation of excessive noise. Overall, this framework injects a controlled amount of sample noise during the learning process and effectively captures multi-modal information, thereby enhancing the model's robustness and performance. Below, we will introduce each component of UGG-ReID with details.

\subsection{Feature Initialization}
To align with prior research~\cite{wang2024demo, Wang2024MambaPro}, we adopt the visual encoder of CLIP with a shared backbone to extract initial features $x^{m} = \{x^{m}_c, x^{m}_p\}$ of multi-modal data. Here, $m\in \{R, N, T\}$ 
indexes the RGB, Near‑Infrared (NIR), and Thermal Infrared (TIR) modalities, while $x^{m}_c$ and $x^{m}_p = \{x^{m}_{1}, x^{m}_{2} \cdots x^{m}_{n}\}$ denote class tokens and local tokens, respectively. $n$ is the number of local tokens.

\subsection{Gaussian Patch-Graph Representation}
The proposed Gaussian Patch-Graph Representation (GPGR) model aims to adopt Gaussian distributions as node representations and then conduct message passing between Gaussian distributions. GPGR mainly contains two key components:  Gaussian Patch-Graph Learning for relationship modeling and Gaussian Patch-Graph Convolutional Network for message passing. Next, we will introduce the above two modules in detail.

\subsubsection{Gaussian Patch-Graph Learning}
In this section, we construct a Gaussian Patch-Graph Learning (GPGL) based on the extracted initial features to capture context-aware dependencies. Let $G(V^{m}, E^{m})$ be a Gaussian patch-graph, $V^{m}$ denotes the node set of the $m$-th modality and $E^{m}$ represents the corresponding edge set.
Different from deterministic learning, we assume that node representation follows a Gaussian distribution $\mathcal{N}\!\left(\boldsymbol{\mu},\,\boldsymbol{\sigma}^{2}\right)$ to distinguish between regions with sufficient and insufficient cues and enhance the ability to learn complex localizations.
To be specific, given the initial global-local feature set ${\mathbf{x}^{m}} = \{{x}^{m}_{c}, {x}^{m}_{1}, {x}^{m}_{2} \cdots {x}^{m}_{n}\} \in \mathbb{R}^{N \times D}$, where $N = n+1$ and $D$ are the number of nodes and the dimension of each node. We first obtain the mean and standard deviation for global and local nodes as follows,
\vspace{-0.3em}
\begin{align}\label{FC}
    {\mu}^{m}_{c} &= \text{FC}_{\mu}({x}^{m}_{c}), \quad {\mu}^{m}_{i} = \text{FC}_{\mu}({x}^{m}_{i}), \quad i=1 \cdots n,
    \\\nonumber
    {\sigma}^{m}_{c} &= \text{FC}_{\sigma}({x}^{m}_{c}), \quad {\sigma}^{m}_{i} = \text{FC}_{\sigma}({x}^{m}_{i}), \quad i=1 \cdots n,
\end{align} 
where $\text{FC}_{\mu}$ and $\text{FC}_{\sigma}$ are projection layers to learn the mean and standard deviation, respectively.
We denote the means and variances by  ${\boldsymbol{\mu}^{m}} = \{{\mu}^{m}_{c}, {\mu}^{m}_{1}, {\mu}^{m}_{2} \cdots {\mu}^{m}_{n}\} \in \mathbb{R}^{N \times D} $  and $\boldsymbol{\sigma}^m = \{{\sigma}_c^m, {\sigma}_1^m, {\sigma}_2^m \cdots {\sigma}_n^m\}\in \mathbb{R}^{N \times D}$, respectively.
%\textbf{Edges: }
Then, we calculate the edge weight in $E^{m}$ and construct the structural relationship to achieve global and local information interaction. 
Let $\boldsymbol{\alpha}^{m}$ be the adjacency matrix and encode structural relationships of $m$-th modality.  Specifically, we compute the similarity by utilizing the mean vectors ${\boldsymbol{\mu}^{m}}$ to avoid the interference of high uncertain nodes, 
\begin{equation}\label{adj}
    {{\alpha}}_{ij}^{m} = \mathrm{Similarity}({{\mu}_i^{m}},{{\mu}_j^{m}}),
\end{equation}
where ${\mu}_i^{m}$ and ${\mu}_j^{m}$ are the mean representations of $i$-th and $j$-th nodes in $m$-th modality. $\mathrm{Similarity}$ is a metric function and we adopt Euclidean distance~\cite{10.1145/3459637.3482261} in our experiments.

\subsubsection{Gaussian Patch-Graph Convolutional Network}
After contructing the Gaussian Patch-Graph, we further employ a Gaussian Patch-Graph Convolutional Network (GPGCN) to facilitate message passing among distributional nodes. Since each node is modeled as a Gaussian distribution, traditional graph convolution operations are not directly applicable. Thus, we adopt GPGCN to separately conduct message passing based on the mean and variance~\cite{10.1145/3292500.3330851} and thus enable effective propagation of both semantic representations and uncertainty information.
Formally, the message passing process is defined as,
\vspace{-0.2em}
\begin{equation}
\begin{aligned}
    \boldsymbol{\hat{\mu}}^{(m,l+1)} &= \mathrm{ReLU}\left[{(\mathcal{D}^{m})}^{-\frac{1}{2}}\boldsymbol{\alpha}^{m}{(\mathcal{D}^{m})}^{-\frac{1}{2}}{\boldsymbol{\hat{\mu}}}^{(m,l)}\boldsymbol{w_{\hat{\mu}}}^{(m,l)}\right], \\
    \boldsymbol{\hat{\sigma}}^{(m,l+1)} &= \mathrm{ReLU}\left[{(\mathcal{D}^{m})}^{-\frac{1}{2}}\boldsymbol{\alpha}^{m}{(\mathcal{D}^{m})}^{-\frac{1}{2}}{\boldsymbol{\hat{\sigma}}}^{(m,l)}\boldsymbol{w_{\hat{\sigma}}}^{(m,l)}\right],
\end{aligned}
\end{equation}
where 
% with diagonal entries $\mathcal{D}^m_{ii}=\sum_{j}{\alpha}_{ij}^m$. 
$\boldsymbol{\hat{\mu}}^{(m,0)}=\boldsymbol{\mu}^{m}$ and $\boldsymbol{\hat{\sigma}}^{(m,0)}=\boldsymbol{\sigma}^{(m)}$. $l=0, 1 \cdots L-1$ denotes the $l$-th layers of GGCN and $\mathcal{D}^{m}$ is corresponding degree matrix of $\boldsymbol{\alpha}^{m}$. 
$\boldsymbol{w_{\hat{\mu}}}^{(m,l)}$ and $\boldsymbol{w_{\hat{\sigma}}}^{(m,l)}$ are trainable weight matrices. 
After stacking multiple layers, we can obtain the final output $\boldsymbol{\hat{\mu}}^{(m, L)}$ and $\boldsymbol{\hat{\sigma}}^{(m, L)}$.
For convenience, let $\boldsymbol{\hat{\mu}}^{m}=\boldsymbol{\hat{\mu}}^{(m, L)}$ and $\boldsymbol{\hat{\sigma}}^{m}=\boldsymbol{\hat{\sigma}}^{(m, L)}$ in the following text. 
Considering that node representations follow Gaussian distributions, we adopt the sampling operation to obtain final output, i.e., 
$\boldsymbol{\hat{x}}^m  \sim \mathcal{N}(\boldsymbol{\hat{\mu}}^m, \boldsymbol{\hat{\sigma}^{2}}^m)$. However, this sampling process is not 
differentiable. Therefore, we employ the reparameterization trick to generate node representations as follows,  
% \vspace{-0.1em}
\begin{equation}\label{adj}
    \boldsymbol{\hat{x}}^m = \boldsymbol{\hat{\mu}}^m + \epsilon \boldsymbol{\hat{\sigma}}^m, \quad \epsilon \sim \mathcal{N}(0, I)
\end{equation}
where $\epsilon$ follows a normal distribution. $\boldsymbol{\hat{x}}^m = \{\hat{x}^m_c, \hat{x}^m_1 \cdots \hat{x}^m_n \}$ represents the final node representations. 
To alleviate the effect of extremely unbalanced standard deviation, we adopt 
a sigmoid function to map $\boldsymbol{\hat{\sigma}}$ to $[0,1]$. %thus reducing the di between different tokens and avoiding overly discrete features that lead to model instability. 
In addition, we also introduce a learnable hyperparameter $\phi$ that can dynamically adjust the constraint strength of the standard deviation according to the data. This allows the model to flexibly cope with different noise levels while maintaining robustness. Finally, to extract a more effective embedding of global information, local cues are aggregated into global messages to enhance features as follows,
\begin{equation}\label{x}
    {\tilde{x}}^m = \big[\hat{x}^m_c, {\eta}[\hat{x}^m_1, \hat{x}^m_2 \cdots \hat{x}^m_n]\big]W^m,
\end{equation}
where $\eta$ denotes a pooling operation. $[\cdot, \cdot]$ is the concatenation operation and $W^m$ is the transformation matrix. To further promote joint aggregation representation, Kullback-Leibler divergence~\cite{9157756} is applied to the class token as,
\vspace{-0.2em}
\begin{equation}\label{KL}
    \mathcal{L}^{m}_{c} = {KL}[\mathcal{N}(x_c^{m} | {{\mu}}^{m}_c, ({{\sigma}}^{m}_c)^{2}) \, \| \, \mathcal{N}(\epsilon | 0, I)]= -\frac{1}{2} \big( 1 + \log({{\sigma}}^{m}_c)^2 - ({{\mu}}^{m}_c)^2 - ({{\sigma}}^{m}_c)^2 \big).
\end{equation}

\subsection{Uncertainty-Guided Mixture of Experts}
Based on the proposed GPGR, we obtain a more abundant feature representation in each modality. To further suppress the possible noise interference in each modality and enhance the multi-modal semantic consistency, we propose the Uncertainty-Guided Mixture of Experts (UGMoE) strategy, which aims to achieve robust and deep multi-modal collaboration by modeling the sample-level uncertainty and guiding the sharing of more expressive experts among modalities.

\subsubsection{Uncertainty-Guided Experts Network}
In the traditional MoE, the selection of experts is often based on the input content. In the application of multi-modal data, experts are usually selected independently in each modality, which makes it easy to ignore the sample noise and complementarity between modalities~\cite{wang2024demo, /nips/YunCPWBZXLC24}. The proposed experts network can bridge this weakness. To be specific,  we first model the uncertainty of the extracted features of each modality to enhance the ability of experts to process samples with different noise levels. The multivariate Gaussian distribution that maps the features of each modality is defined as,
\vspace{-0.1em}
\begin{equation}\label{adj}
    p(z^{m} | \tilde{x}^{m}) \sim \mathcal{N}\big({\tilde{\mu}}^{m}, ({\tilde{\sigma}}^{m})^2\big),
\end{equation}
where the means $\tilde{\mu}$ and variances $\tilde{\sigma}$ are obtained by two independent fully connected layers, respectively. Then, to model the uncertainty, we resample from the distribution as follows,
% \vspace{-0.1em}
\begin{equation}\label{adj}
    {z}^m = \tilde{\mu}^m + {\epsilon} \tilde{\sigma}^m, \quad {\epsilon} \sim \mathcal{N}(0, I).
\end{equation}

However, to ensure the stability and robustness of the model expression, we do not directly use the noisy sampled feature ${z}^m$ in the prediction process of the final task, but use the mean as the final feature representation of the sample for the downstream decision modeling. In addition to this, to further the uncertainty modeling capabilities, we introduce the KL divergence~\cite{9157756} regularization term to ensure that the sample distribution is close to the normal distribution as follows,
\vspace{-0.3em}
\begin{equation}\label{KL}
    \mathcal{L}^{m}_{{s}} = {KL}[\mathcal{N}(z^{m} | {\tilde{\mu}}^{m}, ({\tilde{\sigma}}^{m})^{2}) \, \| \, \mathcal{N}({\epsilon} | 0, I)]= -\frac{1}{2} \big( 1 + \log({\tilde{\sigma}}^{m})^2 - ({\tilde{\mu}}^{m})^2 - ({\tilde{\sigma}}^{m})^2 \big).
\end{equation}
% \vspace{0.4em}
\subsubsection{Uncertainty-Guided Routing}
To better manage input features effectively, we design an uncertainty-guided routing that includes a gate mechanism to gain modal interaction by selecting different modal experts.
The routing process first applies a linear transformation to the input feature. Then, using a softmax activation acts on the result of this transformation to obtain a probability score $S(\tilde{x}^{m})\in \mathbb{R}^{C}$ where $C$ denotes the number of experts. Finally, the ${TOP}_k$ operation is employed to select the top $k$ ($k=1$) excellent experts of the current modality to learn other modalities, optimizing multi-modal interactions.
Thus, each gate involves the total number $C+M-1$ experts, where $M$ denotes the total number of multi-modal data. Besides, to improve the experts' ability, we further add this constraint term~\cite{10.1145/3664647.3680949} as follows,
% \vspace{-0.3em}
\begin{equation}\label{KLr}
    \mathcal{L}^{m}_{{r}} = \frac{1}{C + M - 1} \sum_{c=1}^{C + M - 1} (\tilde{\sigma}_c^{m})^2 S_c^m(\tilde{x}^{m}).
\end{equation}
As $\tilde{\sigma}_c^{m}$ increases, the corresponding expert assigns smaller weights by this constraint.
 
\subsubsection{Interactive Aggregation}
We use gate scores as weights to fuse the expert output results by the uncertainty-guided routing operation above as follows,
\vspace{-0.3em}
\begin{equation}\label{KL}
    \hat{z}^m =\sum_{c=1}^{C + M - 1} S_c (\tilde{x}^{m}) E_c(\tilde{x}^{m}).
\end{equation}

The learned feature tends to be obtained by specific experts, which means that the existence of some experts can not be optimized. To solve this problem, we further add regular terms~\cite{10.1145/3664647.3680949} as,
\vspace{-0.2em}
\begin{equation}
{\mathcal{L}^m_e = \frac{1}{C + M - 1} \sum_{c=1}^{C + M - 1} \left( \frac{1}{B} \sum_{{\tilde{X}}^{m} \in B} {1}\left\{ \arg\max {S_c^m({\tilde{X}}^{m})} = c \right\} \right) \left( \frac{1}{B} \sum_{{\tilde{X}}^{m} \in B} S_c^m({\tilde{X}}^{m}) \right)},
\end{equation}
% \begin{equation}
% L^m_e = \frac{1}{BK} \sum_{\tilde{x}^{m} \in B} \sum_{k=1}^{K + M - 1} {1}\left( \arg\max S_k^m(\tilde{x}^{m}) = k \right) S_k^m(\tilde{x}^{m})
% \end{equation}
where $B$ denotes the batch size and ${\tilde{X}}^{m}$ is the features collection of samples in batch for the $m$-th modality.
The former item refers to the proportion of samples assigned to expert $c$, and the latter item refers to the proportion of weights assigned by the router to expert $c$. Then, we aggregate interactive features via the concatenation operation as  $\mathbf{\hat{z}} = [\hat{z}^R, \hat{z}^N, \hat{z}^T]$.

\subsection{Train Loss}
In the training of the proposed UGG-ReID, we combine multiple loss functions to optimize the overall framework.
First, $\mathcal{L}^{m}_{c,s}$ is used to constrain the global features, which represents the sum of $\mathcal{L}^{m}_{c}$ and $\mathcal{L}^{m}_{s}$. Then, to improve the network of uncertainty-guided experts, we introduce $\mathcal{L}^{m}_{r}$, which prioritizes experts with low uncertainty by dynamically adjusting expert selection, while adopting $\mathcal{L}^{m}_{e}$ to prevent over-reliance on certain experts. Finally, cross-entropy loss $\mathcal{L}_{ce}$ and triplet loss $\mathcal{L}_{tri}$ are used to supervise the entire network. This optimization loss can be expressed by the following formula,
\vspace{-0.2em}
\begin{equation}
\mathcal{L} = \mathcal{L}_{ce} + \mathcal{L}_{tri} + \sum_{m \in \{R, N, T\}}({\lambda}_1\mathcal{L}^{m}_{c,s} + {\lambda}_2\mathcal{L}^{m}_{r} + {\lambda}_3\mathcal{L}^{m}_{e}),
\end{equation}
where ${\lambda}_1$, ${\lambda}_2$ and ${\lambda}_3$ are the balancing coefficients of this overall loss term.

\section{Experiments}
In this section, we evaluate the effectiveness of the proposed UGG-ReID on five commonly used datasets and compare it with some state-of-the-art methods.

\subsection{Experiments Setting}
\label{Setting}
\textbf{Datasets}. We conduct five multi-modal object ReID datasets, including two person ReID datasets (e.g., RGBNT201\cite{Zheng2021RobustMP}, Market1501-MM~\cite{Wang2022InteractEA}) and three vehicle datasets (e.g., MSVR310~\cite{ZHENG2023101901}, RGBNT100~\cite{Li2020MultiSpectralVR}, WMVEID863~\cite{ZHENG2025102800}).
These datasets pose multi-dimensional challenges such as perspective changes and environmental disturbances, reflecting the broad applicability of this method.

\textbf{Implementation Details.}
All experiments are conducted using PyTorch on a single NVIDIA RTX 4090 GPU. A pre-trained CLIP model serves as the visual encoder. Person and vehicle images are resized to 256×128 and 128×256, respectively. We extract features using a 16×16 patching strategy, yielding 128 local tokens and one global token, which jointly serve as nodes in a Gaussian patch-graph for modeling. The model is fine-tuned with Adam (learning rate: 0.00035) for 40 epochs. More details of the experiments are provided in the supplementary material.

\textbf{Evaluation Protocols.} We utilize mAP and Rank to evaluate the performance of the model, where mAP means the accuracy of ReID, while Rank shows the probability that the correct match is included in the top results. The combination of the two can more accurately reflect the ability to identify.

%%%%%%
\begin{table*}[t]
\caption{Comparison with state-of-the-art methods on the multi-modal person ReID datasets(in \%).} 
\label{Person}
\centering
\resizebox{1\columnwidth}{!}{
\begin{tabular}{c|c|c|c|cccc|cccc}
\toprule
&{\multirow{2}{*}{\textbf{Methods}}}
&\multirow{2}{*}{\textbf{Publication}}&\multirow{2}{*}{\textbf{Structure}}
&\multicolumn{4}{c|}{\textbf{RGBNT201}}
&\multicolumn{4}{c}{\textbf{Market1501-MM}}\\
\cmidrule{5-8}
\cmidrule{9-12}
&&&&\textbf{mAP} & \textbf{R-1}&\textbf{R-5} & \textbf{R-10}&\textbf{mAP} & \textbf{R-1}&\textbf{R-5} & \textbf{R-10}\\
\midrule
\multirow{12}{*}{\rotatebox{90}{\textbf{Multi-modal}}}
&{HAMNet}~\cite{Li2020MultiSpectralVR}& {AAAI20}&CNN&27.7&26.3& 41.5&51.7&60.0& 82.8& 92.5 &95.0 \\
&{PFNet}~\cite{Zheng2021RobustMP}& {AAAI21}&CNN&38.5& 38.9& 52.0 &58.4&60.9 &83.6& 92.8& 95.5 \\
&{IEEE}~\cite{Wang2022InteractEA}& {AAAI22}&CNN&
 46.4& 47.1& 58.5& 64.2&64.3& 83.9 &93.0 &95.7 \\
 &{TIENet}~\cite{Yang25TIENet}& {TNNLS25}&CNN&
 54.5& 54.4& 66.3& 71.1&67.4& 86.1 &94.1 &96.0 \\
&{UniCat}~\cite{Crawford2023UniCatCA}& {NIPSW23}&ViT&
 57.0& 55.7& -& -&-& - &- &- \\
&{EDITOR}~\cite{10654953}&{CVPR24}&ViT&66.7&68.7&82.2&87.9&77.4&90.8&96.8&98.3 \\
&{RSCNet}~\cite{10772090}& {TCSVT24}&ViT&68.2 &72.5&-&-&-&-&-&-\\
&{HTT}~\cite{Wang2024HeterogeneousTT}& {AAAI24}&ViT&71.1&73.4&83.1&87.3&67.2& 81.5& 95.8& 97.8 \\
&{TOP-ReID}~\cite{Wang2023TOPReIDMO}& {AAAI24}&ViT&72.2&75.2&84.9&89.4&{82.0}	&{92.4}&	{97.6}&{98.6}\\
&{ICPL-ReID}~\cite{li2025icpl}& {TMM25}&CLIP&{75.1}&77.4&84.2&{87.9}&-&-&-&-\\
&{PromptMA}~\cite{Zhang25PromptMA}& {TIP25}&CLIP&{78.4}&{80.9}&{87.0}&{88.9}&83.6&\underline{93.3}&-&-\\
&{MambaPro}~\cite{Wang2024MambaPro}& {AAAI25}&CLIP&78.9&\underline{83.4}&{89.8}&{91.9}&\underline{84.1}&92.8&\underline{97.7}&98.7\\
&{DeMo}~\cite{wang2024demo}& {AAAI25}&CLIP&{79.7}&81.8&89.4&{92.5}&83.6&{93.1}&{97.5}&\underline{98.7}\\

&{IDEA}~\cite{wang2025IDEA}& {CVPR25}&CLIP&\underline{80.2}&{82.1}&\underline{90.0}&\underline{93.3}&-&-&-&-\\
\midrule
&{\textbf{UGG-ReID}}& \textbf{Ours}&CLIP&
\textbf{81.2}&\textbf{86.8}&\textbf{92.0}&\textbf{94.7}&\textbf{85.4}&\textbf{94.3}&\textbf{ 98.4}&\textbf{99.1}\\
\bottomrule
\end{tabular}
}
\vspace{-2em}
\end{table*}
%%%%%%

\subsection{Comparison with State-of-the-Art Methods}
\textbf{Evaluation on Multi-modal Person ReID}. 
We evaluate our proposed UGG-ReID on two multi-modal person ReID datasets in Table~\ref{Person}. We can observe that both datasets achieve state-of-the-art results for all metrics. 
Particularly, RGBNT201~\cite{Zheng2021RobustMP} outperforms the next most popular method in the metric rank-1 by 3.4\%. 
In addition, our method surpasses DeMo~\cite{wang2024demo} in multiple metrics, highlighting the effectiveness of our architectural design. DeMo~\cite{wang2024demo} relies on modality decoupling to preserve specific cues and uses attention to assign expert weights in MoE. In contrast, our proposed GPGR  builds stronger modality-specific representations by incorporating aleatoric uncertainty from local details. Meanwhile, the UGMoE strategy utilizes sample uncertainty to guide expert selection and applies a novel routing strategy to facilitate more effective multi-modal collaboration.

\textbf{Evaluation on Multi-modal Vehicle ReID}.
We further conduct experiments on three multi-modal vehicle datasets, which contain challenges such as large view discrepancies and intense glare conditions, to fully validate the robustness and effectiveness of the proposed UGG-ReID. As shown in Table~\ref{Vehicle}, our method maintains stable performance under various complex interference conditions, clearly demonstrating the practical effectiveness of the proposed method in enhancing the model’s generalization capability. Notably, on WMVEID863~\cite{ZHENG2025102800} with severe dazzle interference, our method outperforms the suboptimal method by 2.8\% and 3.6\% in the mAP and R-1, respectively. This result further validates the robustness and effectiveness in the face of significant noise interference.

%%%%%%
\begin{table*}[t]
\caption{Comparison with state-of-the-art methods on the  multi-modal vehicle ReID datasets(in \%).} 
\label{Vehicle}
\centering
\resizebox{1\columnwidth}{!}{
\begin{tabular}{c|c|c|c|cc|cc|cccc}
\toprule
&{\multirow{2}{*}{\textbf{Methods}}}
&\multirow{2}{*}{\textbf{Publication}}&\multirow{2}{*}{\textbf{Structure}}
&\multicolumn{2}{c|}{\textbf{MSVR310}}&\multicolumn{2}{c|}{\textbf{RGBNT100}}&\multicolumn{4}{c}{\textbf{WMVEID863}}\\
\cmidrule{5-8}
\cmidrule{9-12}
&&&&\textbf{mAP} & \textbf{R-1}&\textbf{R-5} & \textbf{R-10}&\textbf{mAP} & \textbf{R-1}&\textbf{R-5} & \textbf{R-10}\\
\midrule
\multirow{12}{*}{\rotatebox{90}{\textbf{Multi-modal}}}
&{HAMNet}~\cite{Li2020MultiSpectralVR}& {AAAI20}&CNN&27.1 &42.3& 74.5 &93.3&45.6 & 48.5 & 63.1 & 68.8 \\
&{PFNet}~\cite{Zheng2021RobustMP}& {AAAI21}&CNN&23.5& 37.4& 68.1& 94.1&50.1 & 55.9 & 68.7 & 75.1 \\
&{IEEE}~\cite{Wang2022InteractEA}& {AAAI22}&CNN&21.0 &41.0 &61.3 &87.8&
 45.9 & 48.6 & 64.3 & 67.9 \\
 % &{TIENet}~\cite{Yang25TIENet}& {TNNLS25}&CNN&
 % 54.5& 54.4& 66.3& 71.1&67.4& 86.1 &94.1 &96.0 \\
 &{CCNet}~\cite{ZHENG2023101901}& {INFFUS23}&CNN &36.4 &55.2  &77.2 &96.3&
 50.3 & 52.7 & 69.6 & 75.1 \\
 % 67.4 & 86.1 & 94.1 & 96.0 \\
&{EDITOR}~\cite{10654953}&{CVPR24}&ViT
&39.0 &49.3&82.1& 96.4&65.6 & 73.8 & 80.0 & 82.3 \\
&{RSCNet}~\cite{10772090}& {TCSVT24}&ViT&39.5& 49.6&82.3 &96.6&- &-&-&-\\
&{TOP-ReID}~\cite{Wang2023TOPReIDMO}& {AAAI24}&ViT&35.9 &44.6&	81.2 &96.4&67.7 & 75.3 & 80.8 & 83.5\\
&{FACENeT}~\cite{ZHENG2025102800}& {INFFUS25}&ViT&36.2& 54.1&81.5& 96.9&\underline{69.8} & 77.0 & 81.0 & {84.2} \\
&{PromptMA}~\cite{Zhang25PromptMA}& {TIP25}&CLIP&{55.2}&{64.5}&85.3&97.4&-&-&-&-\\
&{MambaPro}~\cite{Wang2024MambaPro}& {AAAI25}&CLIP&47.0& 56.5&83.9& 94.7&69.5 & 76.9 & 80.6 & 83.8\\
&{DeMo}~\cite{wang2024demo}& {AAAI25}&CLIP
&49.2& 59.8&86.2 &{97.6}&68.8 & \underline{77.2} & \underline{81.5} & 83.8\\
&{IDEA}~\cite{wang2025IDEA}& {CVPR25}&CLIP&47.0 &62.4&\underline{87.2}& 96.5&-&-&-&-\\
&{ICPL-ReID}~\cite{li2025icpl}& {TMM25}&CLIP
&\underline{56.9}& \underline{77.7}&87.0&\textbf{98.6}&67.2 & 74.0 & 81.3 & \underline{85.6}\\
\midrule
&{\textbf{UGG-ReID}}& \textbf{Ours}&CLIP&
\textbf{60.1}&\textbf{78.0}&\textbf{88.0}&\underline{98.1}&\textbf{72.6}&\textbf{80.8}&\textbf{84.2}&\textbf{87.2}\\
\bottomrule
\end{tabular}
}
\vspace{-1.5em}
\end{table*}
%%%%%%
\subsection{Ablation Study}
%%%%
\begin{table*}[t]
\caption{ Ablation study results on the RGBNT201 and WMVEID863 datasets (in \%).}
\label{absolution}
\centering
\resizebox{1\columnwidth}{!}{
\begin{tabular}{ccccccc|cccc|cccc}
\toprule
&\multicolumn{2}{c}{\textbf{UGMoE}} & &\multicolumn{2}{c}{\textbf{GPGR}}& &\multicolumn{4}{c|}{\textbf{RGBNT201}}&\multicolumn{4}{c}{\textbf{WMVEID863}} \\
\cmidrule{2-3}
\cmidrule{5-6}
\cmidrule{8-15}
& \textbf{MoE}& \textbf{Uncer.} && \textbf{PGR} & \textbf{Uncer.} && \textbf{mAP} & \textbf{R-1}& \textbf{R-5} &\textbf{R-10} &\textbf{mAP} &\textbf{R-1}& \textbf{R-5} &\textbf{R-10} \\
\midrule
\textbf{(a)}&$\times$ & $\times$  &&	$\times$& $\times$ & 
&72.2&72.7&82.2&	87.3&66.4&	72.4&	79.4	&82.9
\\
\textbf{(b)}&$\checkmark$ & $\times$   &&	$\times$& $\times$  &
&75.0	&76.3&86.5&89.5&68.7&75.8	&80.2	&82.9
\\	
\textbf{(c)}&$\checkmark$ & $\checkmark$  && $\times$&	$\times$  &
&76.0	&80.4&	88.8&	91.7&69.7&77.0	&81.5&84.5
\\	
\textbf{(d)}&$\checkmark$ & $\checkmark$  && $\checkmark$  & $\times$  &
&77.3&	81.7	&90.3&	92.1
	&70.2&	77.4	&83.5	&86.7
\\	
\midrule
\textbf{(e)}&$\checkmark$ & $\checkmark$  && $\checkmark$  &	$\checkmark$ & 
&\textbf{81.2}&\textbf{86.8}&\textbf{92.0}&	\textbf{94.7}&\textbf{72.6}&\textbf{80.8}&\textbf{84.2}&\textbf{87.2}
\\	

\bottomrule
\end{tabular}
}
\vspace{-1em}
\end{table*}

To analyze the contribution of each module, we conduct systematic ablation experiments around the two core components, UGMoE and GPGR, on RGBNT201~\cite{Zheng2021RobustMP} and WMVEID863~\cite{ZHENG2025102800}. We first use the pre-trained shared CLIP visual encoder as the baseline in Table~\ref{absolution}, and gradually introduce each component for comparative analysis.
\textbf{UGMoE}. Introducing the traditional Mixture of Experts (MoE) strategy to the baseline can bring some performance improvement in Table~\ref{absolution} (b). We further incorporate uncertainty modeling into the MoE strategy and the model performance continues to improve. Compared with the baseline, the proposed method improves mAP/R-1 by 4.8\%/7.7\% and 3.3\%/4.6\% on RGBNT201~\cite{Zheng2021RobustMP} and WMVEID863~\cite{ZHENG2025102800}, respectively, which indicates that the introduction of uncertainty helps to estimate the expert's credibility more accurately, thus realizing a more reasonable sample assignment and improving the effectiveness of multi-modal information fusion.
\textbf{GPGR}. In the case of relying only on the above strategy for multi-modal modeling, the modal-specific information is still not fully explored. To further promote the model's ability, we introduce GPGR to enhance the modeling ability of fine-grained local cues by using uncertainty. In the experiment, we first evaluate the effect of the standard Patch-Graph Representation (PGR) module in Table~\ref{absolution} (d), and then integrate the uncertainty into the network. The results show that GPGR delivers significant improvements based on the integrated UGMoE. Compared to using UGMoE alone, our method improves mAP/R-1 by 5.2\%/6.4\% on RGBNT201~\cite{Zheng2021RobustMP} and by 2.9\%/3.8\% on WMVEID863~\cite{ZHENG2025102800}, respectively. This fully shows that GPGR can effectively suppress noise interference and mine richer information. In summary, our proposed method enhances features within each modality and jointly captures complementary information among multi-modal data, improving model robustness and effectiveness.
More ablation is provided in the supplementary material.

\begin{figure}[!ht]
  \centering
  \includegraphics[width=1\linewidth]{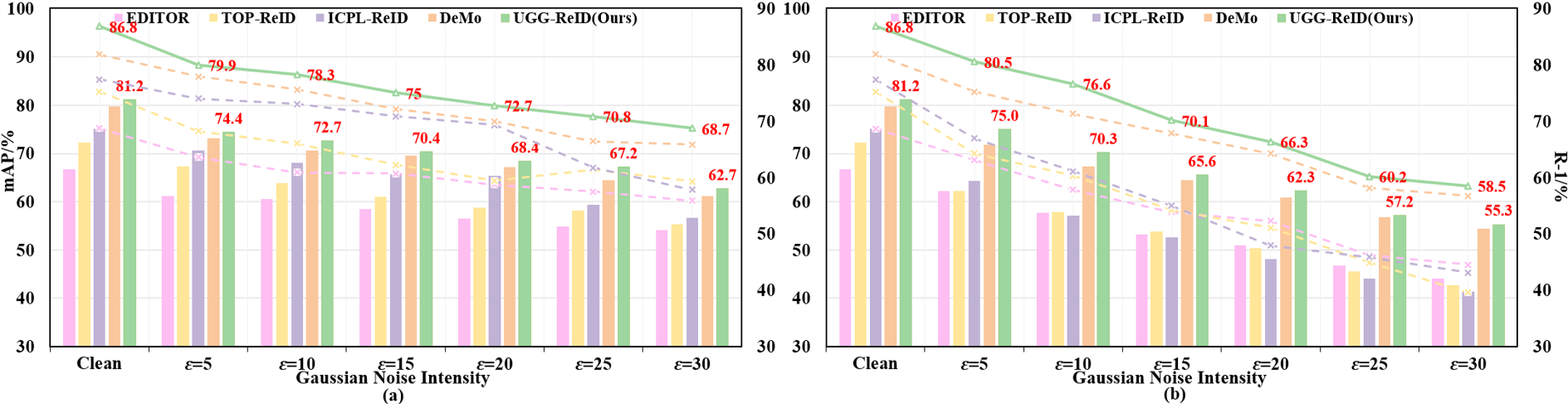} 
  \vspace{-2em}
  \caption{Robustness analysis on RGBNT201. Evaluation results with (a) different levels of Gaussian noise added during dataset generation, and (b) varying noise intensities added during testing after training on clean data.}
  \label{vis2}
  \vspace{-1em}
\end{figure}

\subsection{Robustness Analysis}
To systematically evaluate the robustness of the proposed UGG-ReID under noise interference, we inject Gaussian noise of different intensities into the RGBNT201 dataset~\cite{Zheng2021RobustMP}, and increase the noise intensity $\varepsilon$ from 5 to 30 to generate multiple noisy versions of the dataset. Compared to the four mainstream methods (EDITOR~\cite{10654953}, TOP-ReID~\cite{Wang2023TOPReIDMO}, ICPL-ReID~\cite{li2025icpl} and DeMo~\cite{wang2024demo}), the proposed method maintains superior performance under all noise intensities, as shown in Fig.~\ref{vis2} (a).  These results indicate that the method has good robustness and generalization ability.

\begin{figure}
  \centering
  \includegraphics[width=1\linewidth]{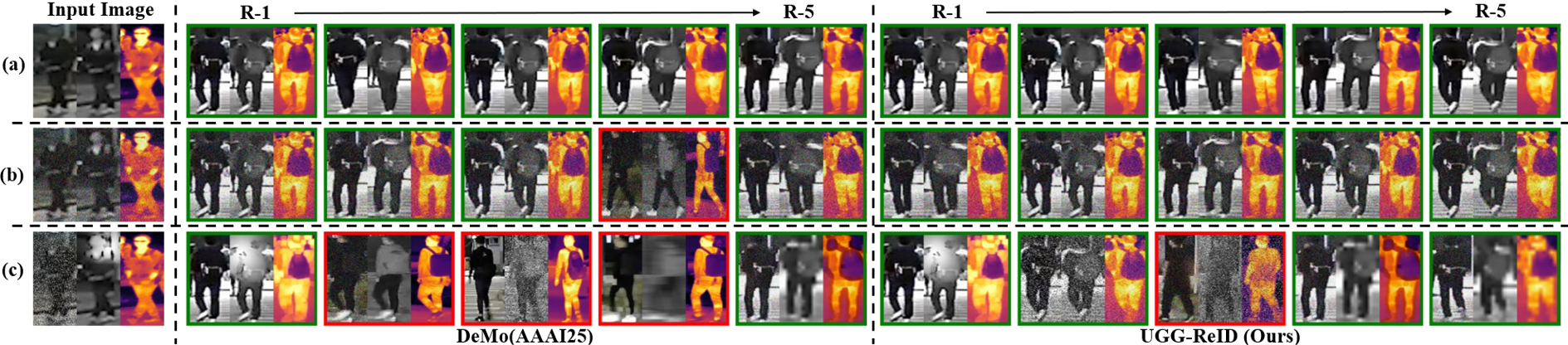} 
  \vspace{-2.0em}
  \caption{Retrieval results under different testing conditions after training on clean data. (a) Clean. (b) Gaussian noise. (c) Arbitrary noise. Green/Red boxes indicate correct/incorrect retrieval results.}
  \label{vis1}
  \vspace{-0.8em}
\end{figure}

Furthermore, we evaluate the model's performance by injecting Gaussian noise of different intensities in the testing phase after completing training on clean data. As Fig.~\ref{vis2} (b) shows, the proposed method is stable and superior to other methods under multi-level noise conditions. Meanwhile, we perform rank-list retrieval evaluations on clean data and different types of noise in Fig.~\ref{vis1}. The results show that the proposed method achieves excellent performance under various interference conditions, which is significantly better than the advanced method DeMo~\cite{wang2024demo}, which fully reflects its robustness and practicability under complex perceptual interference.

\begin{table}
\centering
\begin{minipage}{0.38\linewidth}
    \centering
    \caption{Expert-balanced analysis of each modality on WMVEID863.}
    \label{expert}
    % \vspace{-0.5em}
    \scalebox{0.81}{
    \begin{tabular}{cc|cccc}
        \toprule
         {{\textbf{M}}}&\textbf{Score}& \textbf{\boldmath $E_1$} & \textbf{\boldmath $E_2$} & \textbf{\boldmath $E_3$} & \textbf{\boldmath $E_4$} \\
        \midrule
        \multirow{2}{*}{R} &Uncer.& 0.31&	0.27&	0.30&	0.12\\
         &Gate &0.07&	0.12&	0.07&	0.74\\
         \midrule
        \multirow{2}{*}{N}&Uncer. & 0.24&0.24&0.27&0.24\\
         &Gate&0.28&	0.26	&0.18	&0.28\\
         \midrule
        \multirow{2}{*}{T}&Uncer. &0.28&0.26&0.29&0.18\\
        &Gate  &0.17&	0.21&	0.14&	0.48\\
        \bottomrule
    \end{tabular}}
\end{minipage}
\hfill
\begin{minipage}{0.61\linewidth}
    \caption{ Efficiency analysis and accuracy comparison with SOTA methods on RGBNT201.}
    \label{efficiency}
    % \vspace{-0.5em}
    \centering
    \scalebox{0.81}{
    \begin{tabular}{c|ccccc}
        \toprule
         \textbf{RGBNT201}&\textbf{Params(M)}& \textbf{FLOPs(G)} & \textbf{FPS} & \textbf{mAP} & \textbf{R-1} \\
        \midrule
        TOP-ReID~\cite{Wang2023TOPReIDMO}&324.5&	35.5&	398.9&	72.2&	75.2\\
        EDITOR~\cite{10654953}&119.3&	40.8&	335.1&	66.7&	68.7\\
        PromptMA~\cite{Zhang25PromptMA}&107.9&	36.2&	343.5&	78.4&	80.9\\
        MambaPro~\cite{Wang2024MambaPro}&74.8&	52.4&	243.2&	78.9&	83.4\\
        DeMo~\cite{wang2024demo}&98.8&	35.1&	403.6&	79.7&	81.8\\
        IDEA~\cite{wang2025IDEA}&91.7&	43.7&	299.5&	80.2&	82.1\\
        \textbf{UGG-ReID}&\textbf{103.2}&\textbf{35.0}&\textbf{	371.4}&\textbf{81.2}&\textbf{86.8}\\
        \bottomrule
    \end{tabular}}
\end{minipage}
\vspace{-0.8em}
\end{table}

\subsection{Expert Balanced Analysis}
We show the router average gate scores and uncertainty scores of all test samples for each modality on WMVEID863 in Table~\ref{expert}. We observe that N and T modalities show a more balanced expert activation distribution, while Expert 4 in the R modality has a heavy weight when the uncertainty is low. These results indicate that the routing mechanism dynamically adjusts expert allocation according to different modalities and sample uncertainty.

\subsection{Efficiency Analysis}
To further validate the effectiveness of our proposed UGG-ReID, we conduct an efficiency analysis, as shown in Table~\ref{efficiency}.
We evaluate the inference speed of each method on the RGBNT201 dataset, using Frames Per Second (FPS) as the evaluation metric. As shown in the Table~\ref{efficiency}, UGG-ReID reaches an inference speed of 371.4 FPS while maintaining a relatively low parameter count of 103.2 million and a computational cost of 35.0G FLOPs. This speed is only slightly lower than that of the lighter DeMo~\cite{wang2024demo}, and is significantly higher than most mainstream approaches, including MambaPro~\cite{Wang2024MambaPro} and EDITOR~\cite{10654953}. Notably, despite its high efficiency, UGG-ReID still achieves excellent results.

\begin{figure}[t]
  \centering
  \includegraphics[width=1\linewidth]{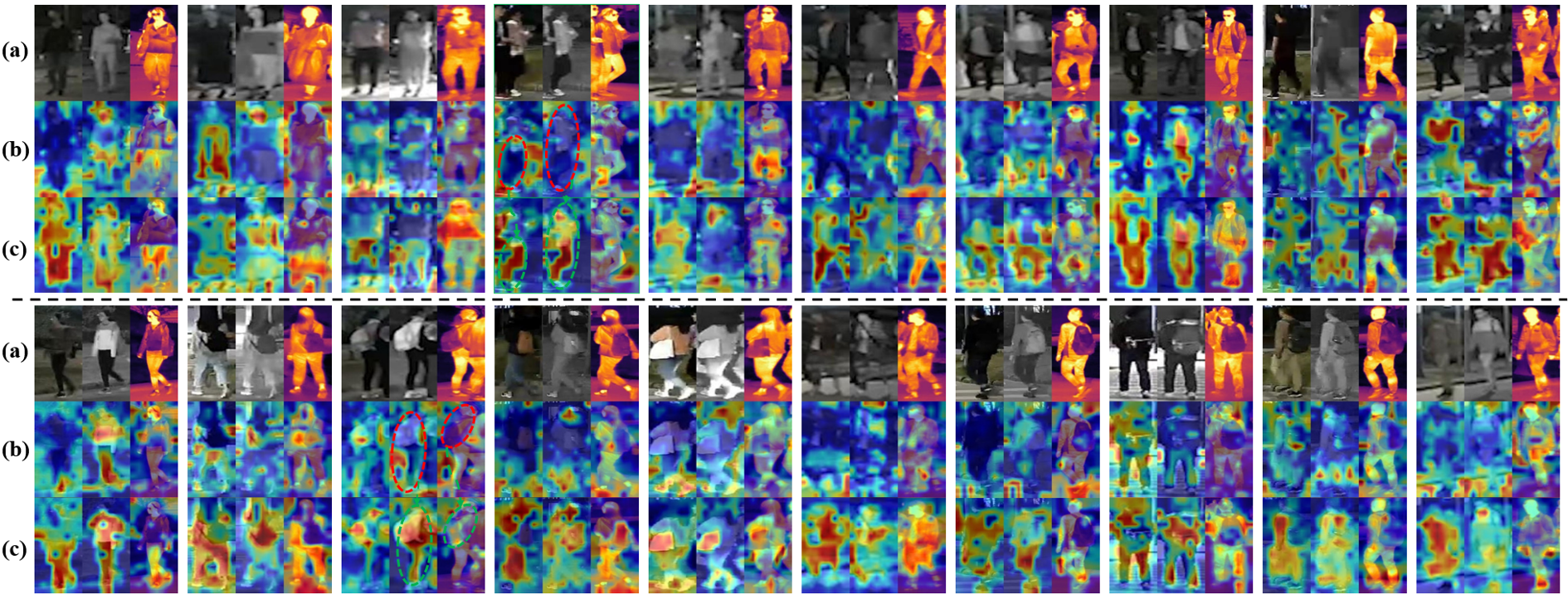} 
  \vspace{-1.8em}
  \caption{Visualization results of the (a) Input image. (b) Baseline. (c) UGG-ReID (Ours).}
  \label{vis6}
  \vspace{-1.0em}
\end{figure}

\begin{figure}
  \centering
\includegraphics[width=1\linewidth]{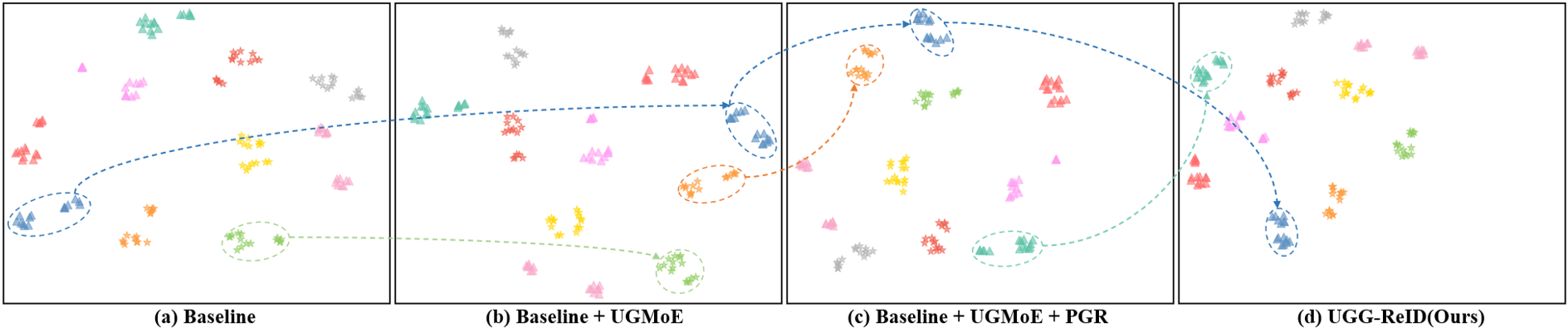}   
  \vspace{-1.5em}
  \caption{T-SNE visualization of extracted features with different model component combinations on the RGBNT201 dataset.}
  \label{vis3}
\vspace{-1.5em}
\end{figure}

\subsection{Visual Results}
\textbf{Multi-modal Activation Maps}: We adopt the Grad-CAM method~\cite{8237336} to visualize the extracted features of each modality. Samples with backpacks are above the dotted line, and samples without backpacks are below the dotted line. The proposed UGG-ReID consistently attends to more semantically rich object regions compared to the baseline model that lacks our module, as shown in Fig.~\ref{vis6}. Notably, under challenging conditions such as motion blur or occlusion, our method still effectively focuses on key target features, further validating its robustness and discriminative power. This suggests that our method effectively captures the diversity of multi-modal information through uncertainty-guided multi-modal local and sample-level joint learning, which significantly improves the robustness and the performance of object re-identification.

\textbf{Multi-modal Feature Distributions}: 
We adopt the T-SNE method~\cite{Kobak2021InitializationIC} to visualize the extracted features to intuitively show the feature distribution of the model under different combinations of modules. With the gradual introduction of the proposed modules, the feature distribution gradually shows obvious clustering in Fig.~\ref{vis3}. This compact clustering structure is consistent with the performance enhancement results in Table~\ref{absolution}, which further validates the effectiveness of our method in enhancing modal feature representation and multi-modal interaction modeling. More visualizations are provided in the supplementary material.

\section{Conclusion}
In this paper, we propose a novel Uncertainty-Guided Graph model for multi-modal object (UGG-ReID), effectively mining modal-specific information and boosting modal collaboration for multi-modal object ReID. 
This UGG-ReID consists of two main aspects. One design is the Gaussian Patch-Graph Representation (GPGR) model for quantifying the aleatoric uncertainty of local and global features and modeling the relationship between them. This manner mitigates the noise interference and enhances the learning ability and robustness of modality-specific information. The other introduces the Uncertainty-Guided Mixture of Experts (UGMoE) strategy to select appropriate experts based on sample uncertainty and facilitate deep multi-modal interactions through a novel routing mechanism.
These two core schemes synergistically optimize the overall network structure and achieve the current optimal performance on all five multi-modal object ReID datasets.

\section*{Acknowledgements}
This work was supported by the National Natural Science Foundation of China (62372003, 62576004), the Natural Science Foundation of Anhui Province (2308085Y40, 2408085J037), the Key Technologies R\&D Program of Anhui Province (202423k09020039), and the Open Research Project of the Anhui Provincial Key Laboratory of Security Artificial Intelligence (SAI202401).

{
\small
\bibliographystyle{neurips_2025}
\bibliography{neurips_2025}
}

\appendix

\clearpage
\section{Supplementary Material}
In the supplementary materials, we provide a detailed description of the UGMoE strategy and present additional experiments to further validate the robustness and effectiveness of the proposed UGG-ReID.

\subsection{Details of UGMoE}
In the main text, we set the value of $k$ in Top $k$ to be 1. In this case, each modality will have $C-1$ unique experts and $M$ shared experts, so the total number of experts is $C + M - 1$.
If $k \neq 1$, we analyze it further to get $C + k(M-1)$ experts for each modality. Thus, Eq.~\ref{KLr} is defined as follows,
\vspace{-0.3em}
\begin{equation}\label{KL}
    \mathcal{L}^{m}_{{r}} = \frac{1}{C + k(M - 1)} \sum_{c=1}^{C +k(M - 1)} (\tilde{\sigma}_c^{m})^2 S_c^m(\tilde{x}^{m}).
\end{equation}
%C+k*(M-1)包含C-K个模态特有专家和KM和共有专家
Next, we utilize gate scores as weights to fuse the expert output results by the above routing operation as follows,
\vspace{-0.3em}
\begin{equation}\label{KL}
    \hat{z}^m =\sum_{c=1}^{C + k(M - 1)} S_c (\tilde{x}^{m}) E_c(\tilde{x}^{m}). 
\end{equation}

The learned feature tends to be obtained by specific experts, which means that the existence of some experts can not be optimized. To solve this problem, we further add regular terms~\cite{10.1145/3664647.3680949} as,
\vspace{-0.3em}
\begin{equation}
{\mathcal{L}^m_e = \frac{1}{C + k(M - 1)} \sum_{c=1}^{C + k(M - 1)} \left( \frac{1}{B} \sum_{{\tilde{X}}^{m} \in B} {1}\left\{ \arg\max {S_c^m({\tilde{X}}^{m})} = c \right\} \right) \left( \frac{1}{B} \sum_{{\tilde{X}}^{m} \in B} S_c^m({\tilde{X}}^{m}) \right)},
\end{equation}

where $B$ denotes the batch size and ${\tilde{X}}^{m}$ is the features collection of samples in batch for the $m$-th modality.
The former item refers to the proportion of samples assigned to expert $c$, and the latter item refers to the proportion of weights assigned by the router to Expert $c$. 
$B$ denotes the batch size. Finally, we aggregate interactive features via the concatenation operation as  $\mathbf{\hat{z}} = [\hat{z}^R, \hat{z}^N, \hat{z}^T]$.

\subsection{Comparison with Prior Works}

For conceptual comparison, we first utilize a Gaussian-based random graph for object representation, where nodes are described by Gaussian distributions to represent the uncertainty of image patches in the presence of noise, whereas previous works generally employ a deterministic graph model, represented by a feature vector. We design a Gaussian Patch-Graph Representation (GPGR) to quantify aleatoric uncertainties for global and local features while modeling their relationships. To our knowledge, this work is the first attempt to exploit a random patch-graph model for the object ReID problem. 
Second, for the Mixture-of-Experts approach, we design the Uncertainty-Guided Mixture of Experts (UGMoE) strategy, which enables different samples to select experts based on uncertainty and utilizes an uncertainty-guided routing mechanism to strengthen the interaction between multi-modal features, effectively promoting modal collaboration.

%%%%
\begin{table*}[b]
\caption{Component-wise comparison of different methods on RGBNT201 (in \%).}
\label{absolution2}
\centering
% \resizebox{1\columnwidth}{!}
{
\begin{tabular}{c|ccccc|cc}
\toprule
\textbf{Method} & \textbf{Uncer.}& \textbf{MoE} &\textbf{Local}& \textbf{Gloabl} & \textbf{Graph} &\textbf{mAP}& \textbf{R-1}  \\
\midrule
EUAR~\cite{10.1145/3664647.3680949}&\checkmark & \checkmark  &$\times$&	\checkmark& $\times$ &74.1 
&77.6
\\
EAU~\cite{10657706}&$\checkmark$ & $\times$   &$\times$&	\checkmark& $\times$  &75.6
&80.3	
\\	
MAP~\cite{10205473}&$\checkmark$ & $\times$  & $\checkmark$&	$\checkmark$  &$\times$ &76.8&78.2
\\	
DeMo~\cite{wang2024demo}&$\times$ & $\checkmark$  &$\checkmark$& $\checkmark$  & $\times$  &79.7&81.8\\	
\midrule
UGG-ReID&$\checkmark$ & $\checkmark$  &$\checkmark$ & $\checkmark$  &	$\checkmark$ & \textbf{81.2}
&\textbf{86.8}
\\	

\bottomrule
\end{tabular}
}
\vspace{-1em}
\end{table*}

For a technical comparison, we first further analyze the impact of different uncertainty modeling approaches on the performance of multi-modal object ReID~\cite{10657706, 10205473, 10.1145/3664647.3680949}. As shown in Table~\ref{absolution2}, works EUAR~\cite{10.1145/3664647.3680949} and EAU~\cite{10657706} are able to perceive inter-sample uncertainty. In contrast, MAP~\cite{10205473} introduces a more comprehensive uncertainty modeling mechanism to quantify the uncertainty of local cues. Although the above methods take uncertainty into account, they either neglect the modeling of uncertainty in local cues or the structural relationships between local regions. Second, we perform comparisons under different MoE strategies. As shown in Table~\ref{absolution2}, we substitute two existing MoE methods~\cite{10.1145/3664647.3680949, wang2024demo}. Compared with DeMo~\cite{wang2024demo}, UGMoE better exploits the diversity of samples by introducing uncertainty modeling, and compared with EUAR~\cite{10.1145/3664647.3680949}, UGMoE further strengthens the interaction between different modalities.

\subsection{Details of Experiments}

\subsubsection{Experiments Setting}
\label{Appendices1}
\textbf{Datasets.} To comprehensively evaluate the generalization ability of the proposed UGG-ReID framework, we conduct experiments on five public datasets. These include two person re-identification datasets, RGBNT201~\cite{Zheng2021RobustMP} and Market1501-MM~\cite{Wang2022InteractEA}, as well as three challenging vehicle re-identification benchmarks: MSVR310~\cite{ZHENG2023101901}, RGBNT100~\cite{Li2020MultiSpectralVR}, and WMVEID863~\cite{ZHENG2025102800}. These datasets collectively reflect a wide range of real-world scenarios and associated challenges. Table~\ref{Dataset} summarizes the partition protocols and the specific challenges posed by each dataset.
\begin{table}[h]
    \centering
    \caption{Details of the datasets partition settings and their corresponding challenges, */* represents ID/Sample.}
    \label{Dataset}
    \resizebox{1\columnwidth}{!}{
    \begin{tabular}{c|c|c|c|c|c}
    \toprule
     & \textbf{RGBNT201}   & \textbf{Market1501-MM}   & \textbf{MSVR310}   & \textbf{RGBNT100}   & \textbf{WMVEID863}   \\
    \midrule
    \textbf{Train}& 171/3951 & 751/12936  & 155/1032   & 50/8675   & 603/10446   \\ 
    \textbf{Query}& 30/836  & 750/3368   & 52/591  & 50/1715  & 210/2904 \\ 
    \textbf{Gallery}& 30/836   & 751/15913  & 155/1055  & 50/8575  & 272/3678  \\ 
    \midrule
    \multirow{2}{*}{\textbf{Challenges}} &Wide Views,  & Simulate the  & Longer Time Span,  & Different Views, & \multirow{2}{*}{Intense Flare}  \\ 
     & Occlusions  & Night Scene & Complex Conditions & Illumination Issue  &   \\ 
    \bottomrule
    \end{tabular}
    }
\end{table}

\textbf{Implementation Details. }
For all experiments, we set  the number of experts at $C=4$
and utilize $k=1$ for the $TOP_k$ selection. The loss terms are weighted with ${\lambda}_1 = 0.1$ and ${\lambda}_2,{\lambda}_3 = 0.0001$, respectively. The number of layers for GPGCN $L$ is set to 2. Our code is implemented in Python using the PyTorch framework and will be released publicly upon acceptance.

\begin{table*}[b]
\caption{ Ablation results for different loss  on the RGBNT201 and WMVEID863 datasets (in \%).}
\label{absolution1}
\centering
% \resizebox{0.8\columnwidth}{!}
{
\begin{tabular}{cc|cccc|cccc}
\toprule
&Loss &\multicolumn{4}{c|}{\textbf{RGBNT201}}&\multicolumn{4}{c}{\textbf{WMVEID863}} \\
\cmidrule{3-10}
&Type& \textbf{mAP} & \textbf{R-1}& \textbf{R-5} &\textbf{R-10} &\textbf{mAP} &\textbf{R-1}& \textbf{R-5} &\textbf{R-10} \\
\midrule
\textbf{(a)}&{w/o {\textbf{$\mathcal{L}_{c,s}$}}}   
&78.8&84.8&89.6&91.7&69.6&76.9&	82.4&85.8\\
\textbf{(b)}&{w/o {\textbf{$\mathcal{L}_{r}$}}} 
&79.0&83.7&91.6&94.1&70.6&79.2&	84.2&86.8\\

\textbf{(c)}&{w/o {\textbf{$\mathcal{L}_{e}$}}} & 
81.0&84.6&90.7&92.5&71.2&78.3&	84.5&87.7\\
% \textbf{(a)}&{w/o {\textbf{$\mathcal{L}_{s}$}}} & 
% 72.2&72.7&82.2&87.3&-&-&	-&-\\

\midrule
\textbf{(d)}& UGG-ReID  
&\textbf{81.2}&\textbf{86.8}&\textbf{92.0}&	\textbf{94.7}&\textbf{72.6}&\textbf{80.8}&\textbf{84.2}&\textbf{87.2}\\
\bottomrule
\end{tabular}
}
\end{table*}

\subsubsection{Ablation Analysis}
To verify the role of each loss in the model, we conduct systematic ablation experiments, as shown in Table~\ref{absolution1}. $\mathcal{L}_{c,s}$ represents the sum of $\mathcal{L}_{c}$ and $\mathcal{L}_{s}$, which is used to impose constraints on the global token. From the experimental results, one can observe that when the $\mathcal{L}_{c,s}$  constraint on the global token is removed, the performance of the model on multiple evaluation indicators decreases, indicating that the constraint has a positive effect on improving modeling ability.
Then, $\mathcal{L}^{m}_{r}$ is removed to verify performance for adding the loss constraint on the expert. We can find that adding the loss, our mAP/R-1 increases by 2.2\%/3.1\% and 2.0\%/1.6\% in RGBNT201~\cite{Zheng2021RobustMP} and WMVEID863~\cite{ZHENG2025102800}, respectively, which verifies its enhancement effect on the expert selection strategy
Finally, we verify the effectiveness of $\mathcal{L}^{m}_{e}$, which aims to ensure that the number of similar samples assigned to each expert in the training process is balanced. Meanwhile, the expert weights are relatively evenly distributed among the experts, and the experimental results show that it can effectively prevent the imbalance of distribution among experts and improve the ability of the model.

\subsubsection{Hyperparameter Analysis}
We analyze the effects of the hyperparameters $C$ and $k$ on model performance, where $C$ controls the number of experts and $k$ denotes the number of shared experts selected for each modality. As shown in Table~\ref{hyperparameters}, a moderate increase in $C$ enhances the model’s expressive capacity, while an appropriate choice of $k$ strikes a balance between stability and flexibility. This facilitates dynamic collaboration and complementarity among experts, ultimately improving overall model performance.

We further analyze the local nodes $n$ of GPGL and Layers $L$ of GPGCN of the GPGR for the effect of the model in the RGBNT201 dataset in Table~\ref{hyperparameters1}. For nodes n, we observe that n=128 achieves excellent results. Too few nodes are not enough to cover rich local information, and too many introduce redundancy and noise, interfering with graph structure learning. For layers $L$, GPGCN works best when $L$=2. The number of layers is too shallow and may lead to insufficient fusion of local structures, while too deep may cause over-smoothing, resulting in the loss of discriminative representation of nodes and weakening the expression ability of local discriminative features.

\begin{table}
\caption{Results of the analysis on hyperparameters $C$ and $k$ on the RGBNT201 and WMVEID863 datasets (in \%).}
\label{hyperparameters}
\centering
\begin{minipage}{0.45\linewidth}
    \centering
    \scalebox{1}{
    \begin{tabular}{c|cc|cc}
        \toprule
         {\multirow{2}{*}{\textbf{Expert $C$}}}& \multicolumn{2}{c|}{\textbf{RGBNT201}} & \multicolumn{2}{c}{\textbf{WMVEID863}}\\
        \cmidrule{2-3}
        \cmidrule{4-5}
         & \textbf{mAP} & \textbf{R-1} & \textbf{mAP} & \textbf{R-1} \\
        \midrule
        2 & 80.3 & 83.7 &71.4 &78.8\\
        3 & 80.8& 85.2& 72.0&80.1\\
        \textbf{4} &\textbf{81.2}&\textbf{86.8}&\textbf{72.6}& \textbf{80.8}\\
        5 &80.0& 84.6& 72.0& 79.7\\
        6 &80.1 &83.7 & 71.1&78.1\\
        % 7 &78.1 &81.5 & 70.7&78.1\\
        % 8 &76.4 &78.7 &70.0 & 77.8\\
        \bottomrule
    \end{tabular}}
\end{minipage}
\hfill
\begin{minipage}{0.45\linewidth}
    \centering
    \scalebox{1}{
    \begin{tabular}{c|cc|cc}
        \toprule
         {\multirow{2}{*}{\textbf{Top $k$}}}& \multicolumn{2}{c|}{\textbf{RGBNT201}} & \multicolumn{2}{c}{\textbf{WMVEID863}}\\

        \cmidrule{2-3}
        \cmidrule{4-5}
         & \textbf{mAP} & \textbf{R-1} & \textbf{mAP} & \textbf{R-1} \\
        \midrule
        0 &78.3 &83.1 &71.3&78.8\\
        \textbf{1} & \textbf{81.2}& \textbf{86.8} & \textbf{72.6}& \textbf{80.8}\\
        2 &79.9 &84.6 &71.7 &79.2\\
        % 3 &79.6 &84.1 & 71.2&78.5\\
        3 &80.2 &85.3 & 71.2&78.5\\
        4 &79.0 &81.6 & 70.3&77.6\\
        \bottomrule
    \end{tabular}}
\end{minipage}
\end{table}

\begin{table}
\caption{Results of the analysis on hyperparameters $L$ and $n$ on the RGBNT201 dataset (in \%).}
\label{hyperparameters1}
\centering
\begin{minipage}{0.45\linewidth}
    \centering
    \scalebox{1}{
    \begin{tabular}{c|cccc}
        \toprule
         {\multirow{2}{*}{\textbf{Layers $L$}}}& \multicolumn{4}{c}{\textbf{RGBNT201}} \\
        \cmidrule{2-3}
        \cmidrule{4-5}
         & \textbf{mAP} & \textbf{R-1} & \textbf{mAP} & \textbf{R-1} \\
        \midrule
        1 & 80.1&	85.3&	91.3	&93.9\\
        \textbf{2} &\textbf{81.2} &	\textbf{86.8} &	\textbf{92.0} 	&\textbf{94.7}\\
        3 & 79.1&	84.6	&90.8	&93.3\\
        4 &78.4	&82.5&	90.3&	92.8\\
        5 &76.3&	80.7&	89.1&	91.7\\
        \bottomrule
    \end{tabular}}
\end{minipage}
\hfill
\begin{minipage}{0.45\linewidth}
    \centering
    \scalebox{1}{
    \begin{tabular}{c|cccc}
        \toprule
         {\multirow{2}{*}{\textbf{Nodes $n$}}}& \multicolumn{4}{c}{\textbf{RGBNT201}}\\

        \cmidrule{2-3}
        \cmidrule{4-5}
         & \textbf{mAP} & \textbf{R-1} & \textbf{mAP} & \textbf{R-1} \\
        \midrule
        32 &79.3&	83.7&	91.3&	93.9\\
        64 & 80.1&	82.9&	90.0	&94.0\\
        96 &\textbf{81.4}&	84.6&	90.4&	92.6\\
        \textbf{128} &{81.2} &\textbf{86.8} &\textbf{92.0}& \textbf{94.7}\\
        160 &79.8&	83.6&	90.2&	91.9\\
        \bottomrule
    \end{tabular}}
\end{minipage}
\end{table}

\begin{figure}[t]
  \centering
  \includegraphics[width=1\linewidth]{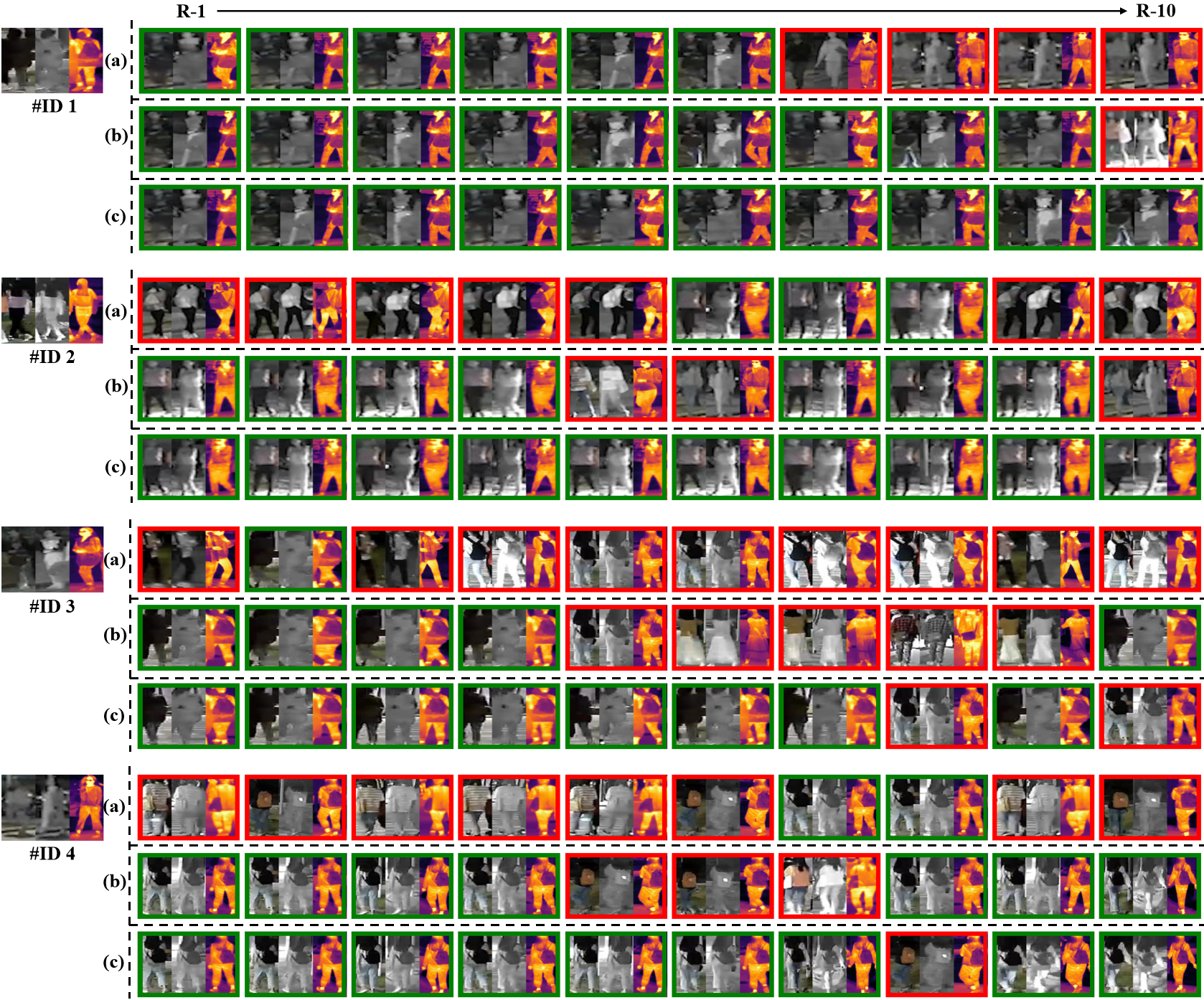}   
  \vspace{-1.8em}
  \caption{Rank-list visualizations for four persons from the RGBNT201 dataset under different model configurations: (a) Baseline, (b) Baseline + UGMoE, and (c) UGG-ReID (Ours).}
  \label{vis4}
\end{figure}
\begin{figure}[t]
  \centering
  \includegraphics[width=1\linewidth]{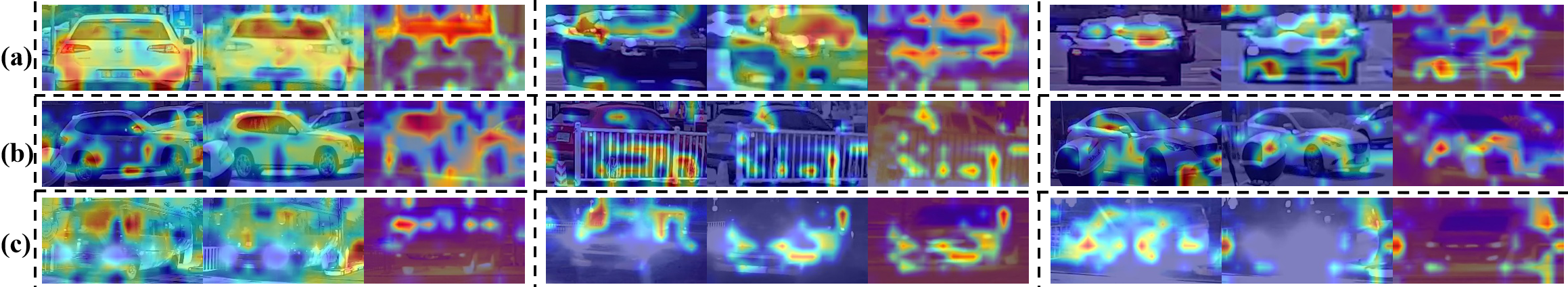}   
  \caption{Visualization of Class Activation Maps (CAMs) under different environmental conditions for six vehicles from the WMVEID863 dataset: (a) Normal, (b) Occlusion, and (c) Intense Flare.}
  \label{vis5}
\end{figure}

\subsubsection{Visual Results}

\textbf{Visualization of Rank List.}
To analyze the performance of the proposed UGG-ReID method in cross-camera retrieval scenarios, we perform rank-list visualization of the retrieval results of different methods as Fig.~\ref{vis4}. Compared with baseline and baseline+UMoE, UGG-ReID can rank the objects more accurately, demonstrating stronger model robustness and discriminative ability.

\textbf{Visualization of Class Activation Maps.} As shown in Fig.~\ref{vis5}, we visualize the proposed UGG-ReID using Class Activation Maps (CAMs)~\cite{8237336}. The results further demonstrate that our approach is capable of capturing discriminative local regions, even under complex environmental conditions.

\subsection{Discussion}
Multi-modal object ReID exploits fine-grained local cues and the complementary information of modalities to effectively enhance the robustness and accuracy of recognition in complex scenarios~\cite{Zheng2021RobustMP,Wang2023TOPReIDMO,wang2025IDEA,Zhang25PromptMA,wang2024demo}. 
As is well known, significant distributional differences exist among different modalities, and noise arising from sample quality and environmental factors further impacts the accuracy of feature representations.
The proposed UGG-ReID effectively guides the feature fusion process by explicitly quantifying local and sample-level epistemic uncertainties and modeling the relationship between them, enhancing the model's robustness and effectiveness.
UGG-ReID is the first work that leverages uncertainty to quantify fine-grained-local details and explicitly model their dependencies in multi-modal data.

\textbf{Limitations and In the Future.}
 Our framework employs uncertainty-guided learning to enhance robustness against local noise; it may still struggle under extreme conditions where local cues are heavily corrupted or missing. In future work,
 we will focus on advancing uncertainty quantification and reasoning techniques, exploring the integration of Bayesian inference and evidence theory into multi-modal object ReID~\cite{Kaushik_2024_CVPR, DCEL, UGNCL}. This aims to enhance the model’s robustness to modality and label noise, thereby improving its overall performance and reliability in complex environments.
\label{Appendices3}

% \section*{References}

% \medskip

% {
% \small

% [39] Alexander, J.A.\ \& Mozer, M.C.\ (1995) Template-based algorithms for
% connectionist rule extraction. In G.\ Tesauro, D.S.\ Touretzky and T.K.\ Leen
% (eds.), {\it Advances in Neural Information Processing Systems 7},
% pp.\ 609--616. Cambridge, MA: MIT Press.

% [40] Bower, J.M.\ \& Beeman, D.\ (1995) {\it The Book of GENESIS: Exploring
%   Realistic Neural Models with the GEneral NEural SImulation System.}  New York:
% TELOS/Springer--Verlag.

% [41] Hasselmo, M.E., Schnell, E.\ \& Barkai, E.\ (1995) Dynamics of learning and
% recall at excitatory recurrent synapses and cholinergic modulation in rat
% hippocampal region CA3. {\it Journal of Neuroscience} {\bf 15}(7):5249-5262.
% }

%%%%%%%%%%%%%%%%%%%%%%%%%%%%%%%%%%%%%%%%%%%%%%%%%%%%%%%%%%%%

\end{document}